\documentclass[journal,twoside]{IEEEtran}
\usepackage[T1]{fontenc}
\usepackage{cite}
\usepackage{amsmath,amssymb,amsfonts}
\usepackage{algorithmic}
\usepackage[]{graphicx}
\usepackage[mathlines,switch]{lineno}
\usepackage{multicol,multirow}
\usepackage{epsfig}
\usepackage{TUSON}
\usepackage{booktabs}
\usepackage{tabularx}
\usepackage{xcolor}
\usepackage{colortbl}
\usepackage{orcidlink}
\usepackage{textcomp}
\usepackage[T1]{fontenc}
\usepackage{booktabs}   % for \toprule, \midrule, \bottomrule
\usepackage{multirow}   % for multi-row cells (used in header)
\usepackage{caption}    % recommended for table captions
\usepackage{amsmath,amssymb}
\usepackage{subcaption}
\usepackage{pifont}
\usepackage{tabularx}
\usepackage{array}
\usepackage{adjustbox}
\newcommand{\ann}[1]{\textsuperscript{(#1)}}
\newcolumntype{Y}{>{\raggedright\arraybackslash}X}
\newcommand{\na}{--}
\usepackage{multirow}
\usepackage{makecell}
\usepackage{booktabs}
\usepackage{graphicx,verbatim}
\usepackage{xcolor}
\usepackage{hyperref}  
\usepackage[capitalise,noabbrev]{cleveref}

\hypersetup{
    colorlinks=true,
    linkcolor=blue,
    filecolor=cyan,      
    urlcolor=magenta,
}
\newcommand{\best}[1]{\textcolor{blue}{#1}}
\newcommand{\second}[1]{\textcolor{red}{#1}}

\title{Prompt2Echo (P2Echo): Text-Prompted Multi-View Echocardiography Segmentation with One Model}
\author{%
Moein Heidari\textsuperscript{\orcidlink{0000-0003-3676-0127}\,*},\;
Alvin Kimbowa\textsuperscript{\orcidlink{0000-0002-7588-5902}\,*},\;
Mahdi Tabatabaei\textsuperscript{\orcidlink{0009-0008-8406-6803}},\;
Afshin Bozorgpour\textsuperscript{\orcidlink{0000-0003-1857-1058}},\;
Sina Ghorbani Kolahi\textsuperscript{\orcidlink{0009-0002-4519-2635}},\;
River Jiang\textsuperscript{\orcidlink{0000-0002-1702-8873}},\;
Wenjin Chen,\;
David~J. Foran\textsuperscript{\orcidlink{0000-0002-0090-0055}},~\IEEEmembership{Member, IEEE},\;
Jasmine Grewal,\;
Dorit Merhof\textsuperscript{\orcidlink{0000-0002-1672-2185}},~\IEEEmembership{Member, IEEE},\;
and Ilker Hacihaliloglu\textsuperscript{\orcidlink{0000-0003-3232-8193}},~\IEEEmembership{Member, IEEE}%
\thanks{Manuscript submitted on <September, 12, 2026>. $^{*}$Equal contribution. (Corresponding author: Moein Heidari).}
\thanks{M. Heidari is with the School of Biomedical Engineering, The University of British Columbia, Vancouver, BC V6T 1Z4, Canada (e-mail: \href{mailto:moein.heidari@ubc.ca}{moein.heidari@ubc.ca}).}%
\thanks{A. Kimbowa is with the School of Biomedical Engineering, The University of British Columbia, Vancouver, BC V6T 1Z4, Canada (e-mail: \href{mailto:alvinbk@student.ubc.ca}{alvinbk@student.ubc.ca}).}%
\thanks{M. Tabatabaei is with École polytechnique fédérale de Lausanne (EPFL), Lausanne, Switzerland (e-mail: \href{mailto:Seyedmahdi.tabatabaei@epfl.ch}{Seyedmahdi.tabatabaei@epfl.ch}).}%
\thanks{A. Bozorgpour is with the Faculty of Informatics and Data Science, University of Regensburg, 93053 Regensburg, Germany (e-mail: \href{mailto:afshin.bozorgpour@ur.de}{afshin.bozorgpour@ur.de}).}%
\thanks{S. Ghorbani Kolahi is an independent computer science researcher, Tehran, Iran (e-mail: \href{mailto:sinaghorbani388@gmail.com}{sinaghorbani388@gmail.com}).}%
\thanks{R. Jiang is with the Division of Cardiology, The University of British Columbia, Vancouver, BC, Canada, and the University of California, San Francisco, CA, USA (e-mail: \href{mailto:riverjiang@me.com}{riverjiang@me.com}).}%
\thanks{W. Chen is with the Biomedical Informatics Division, Rutgers Cancer Institute, Rutgers University, New Brunswick, NJ 08901 USA (e-mail: \href{mailto:chenwe@rwjms.rutgers.edu}{chenwe@rwjms.rutgers.edu}).}%
\thanks{D.~J. Foran is with the Rutgers Cancer Institute and the Departments of Pathology, Laboratory Medicine and Radiology, Rutgers Robert Wood Johnson Medical School, Rutgers University, New Brunswick/Piscataway, NJ 08901 USA (e-mail: \href{mailto:foran@cinj.rutgers.edu}{foran@cinj.rutgers.edu}).}%
\thanks{J. Grewal is with the Division of Cardiology, University of British Columbia, and St. Paul’s Hospital (Provincial Cardio-Obstetrics and Virani Provincial Adult Congenital Heart Programs), Vancouver, BC, Canada (e-mail: \href{mailto:jasmine.grewal@phc.ca}{jasmine.grewal@phc.ca}).}%
\thanks{D. Merhof is with the Chair of Image Analysis and Computer Vision, Faculty of Informatics and Data Science, University of Regensburg, 93053 Regensburg, Germany (e-mail: \href{mailto:dorit.merhof@informatik.uni-regensburg.de}{dorit.merhof@informatik.uni-regensburg.de}).}%
\thanks{I. Hacihaliloglu is with the Department of Radiology and the Department of Medicine, The University of British Columbia, Vancouver, BC V6T 1Z4, Canada (e-mail: \href{mailto:ilker.hacihaliloglu@ubc.ca}{ilker.hacihaliloglu@ubc.ca}).}%
}

\IEEEaftertitletext{\GA{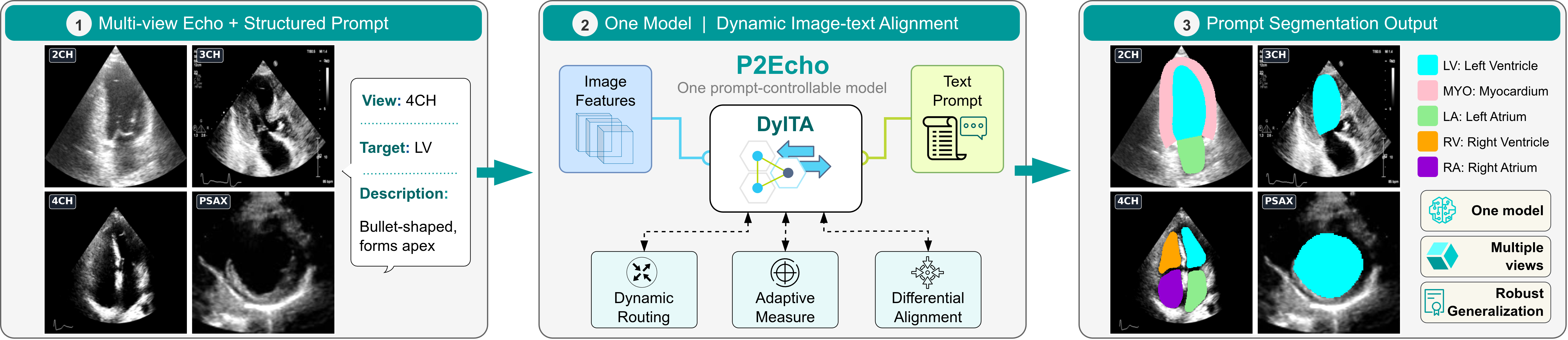}
\Abstract{\begin{abstract}
Automated segmentation in multi-view echocardiography is essential for comprehensive cardiac assessment, yet remains challenging due to substantial anatomical and appearance variability across standard acoustic windows. In this paper, we present P2Echo, a unified framework capable of segmenting diverse cardiac structures across multiple standard echocardiographic views using a single model controlled by structured view-aware text prompts. Unlike recent foundation model adaptations that rely on implicit latent priors, P2Echo explicitly conditions the segmentation process via a novel Dynamic Image-Text Alignment (DyITA) module. DyITA integrates a differentiable routing mechanism and token differential operators to dynamically modulate decoder features according to the requested view and target structure. This allows the network to adaptively switch its processing pathways for different views without architectural changes. Extensive experiments on a large-scale internal dataset (seen during training) and held-out evaluation on three external datasets demonstrate that P2Echo matches or exceeds both view-specific specialists and state-of-the-art universal models at a fraction of their computational cost, offering an efficient, controllable solution for multi-plane cardiac analysis. Code is available at:
\url{https://github.com/moeinheidari7829/P2Echo}.
\hblne
\end{abstract}}
\vspace{1\baselineskip}\vspace*{-1pt}
}

\begin{document}

\maketitle

\Highlights[1]{P2Echo is a single, language-prompted model that segments diverse cardiac structures across echo views, using a DyITA module to align text and image features inside each decoder stage.}

\Highlights[2]{Across five internal and three external datasets, P2Echo matches or exceeds view-specific specialists and foundation models while running at a fraction of their compute (6.90 GFLOPs, 64.68 FPS), with a median gain of 2.33\% Dice across evaluated datasets over the strongest baseline.}

\Highlights[3]{A single prompt-controllable network replaces multiple view-specific specialists, cutting development and maintenance costs and enabling consistent, efficient multi-plane cardiac assessment.}


\PrintHighlights

\section{Introduction}
\label{sec:introduction}

\begin{figure}[!t]
    \centering
    \includegraphics[width=\columnwidth]{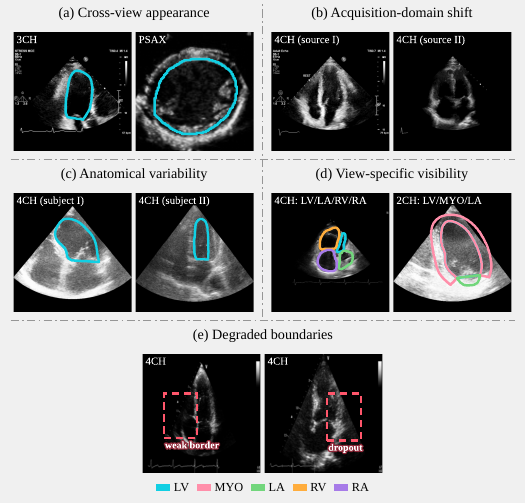}
    \caption{Key challenges in multi-view echocardiography image segmentation: (a) cross-view appearance disparity, (b) acquisition-domain shift, (c) inter-subject anatomical variability, (d) view-specific anatomical visibility, and (e) boundary degradation from speckle, shadowing, and signal dropout. Colored contours denote the annotated cardiac structures.}
    \label{fig:image_challenges}
\end{figure}

Echocardiography is the primary non-invasive modality for cardiac diagnosis, relying on a standard set of acoustic windows (e.g., the Apical 2-, 3-, and 4-Chamber (A2C/2CH, A3C/3CH, A4C/4CH) and Parasternal Long/Short Axis (PLAX, PSAX) views) to visualize complex anatomy \cite{hu2025echoone}. Although multi-view echocardiographic examinations provide complementary information, manual interpretation is time-consuming and prone to inter-observer variability. To enable objective and reproducible quantification of cardiac structure and function, automated analysis is essential, with accurate cardiac segmentation serving as a key foundational step. However, current automated solutions are largely siloed due to the dramatic anatomical appearance shifts between views; traditional deep learning models fail to generalize across planes \cite{li2019recurrent}. This necessitates training separate specialist models for each view \cite{leclerc2019deep}, leading to redundant development, high maintenance costs, and inconsistent performance across the clinical workflow. As summarized in \cref{fig:image_challenges}, multi-view echocardiography segmentation is complicated by several interacting sources of variability. First, the appearance and geometry of the same cardiac structure can change substantially across acoustic windows \cite{hu2025echoone,leclerc2019deep}, while only a subset of structures is visible in each view. Second, differences in scanners, acquisition settings, and clinical centers introduce additional domain shifts that alter image appearance even within the same view \cite{yang2023graphecho}. Inter-subject anatomical variability further changes chamber size, shape, and relative position. Finally, the characteristic speckle pattern of ultrasound, together with acoustic shadowing and signal dropout, can substantially weaken anatomical boundaries. Consequently, a unified model must simultaneously accommodate view-conditioned anatomy, heterogeneous acquisition domains, and locally ambiguous boundaries rather than learning a single view-agnostic appearance representation.

To address this redundancy, recent research has moved toward unified, multi-view segmentation. Early attempts utilized multi-branch architectures, where dedicated encoders process specific views before fusing information through cross-view dependency modeling \cite{ding2022mvfusfra,jabbar2022sala}. While effective, these designs increase model complexity and remain architecturally rigid, as adding a new view requires adding a new parallel branch. Alternatively, universal models attempt to learn a single view-agnostic representation from pooled data. However, empirical evidence shows that when classic architectures are trained on combined multi-plane data, their performance significantly decreases compared to plane-specific training. Without explicit conditioning, these models often converge to an average solution that degrades per-view accuracy, failing to capture the unique geometric details of conflicting views (e.g., the circular left ventricle in short-axis vs. the bullet-shaped ventricle in apical views) \cite{hu2025echoone}.
The rise of Foundation Models has introduced new paradigms for universal segmentation. SAM-based approaches, such as SAMUS \cite{lin2024beyond} and EchoONE \cite{hu2025echoone}, adapt the Segment Anything Model (SAM) \cite{kirillov2023segment} for ultrasound by incorporating adapter layers or latent prompt generators. While EchoONE achieves multi-view capabilities, its dense prompts are derived from clustering-based structural prototypes, making the resulting guidance dependent on how well these learned priors represent the input anatomy and difficult to control explicitly. Parallel to this, text-guided segmentation methods have shown that natural language can serve as an explicit control signal, from MedCLIP-SAMv2 \cite{koleilat2025medclip} to Grounding DINO-US-SAM \cite{rasaee2025grounding}, which pairs a LoRA-tuned Grounding DINO detector with SAM2 to segment multiple organs across 18 ultrasound datasets from free-form prompts. In these methods, language primarily guides target localization or visual-prompt generation before mask decoding, failing to continuously modulate decoder features at multiple spatial resolutions, as required to resolve the cross-view appearance variance of the same anatomical class in multi-view echocardiography.

In this work, we propose P2Echo (Prompt2Echo), a unified framework that leverages structured textual prompts to explicitly condition and control a single segmentation model across diverse echocardiographic views. Unlike prototype-derived latent priors, which are learned implicitly and cannot be directly specified by users, P2Echo employs view-specific text descriptions (e.g., ``Apical 4 Chamber Left Ventricle'') as explicit conditioning signals. The operator can directly define the desired view and target structure, while P2Echo dynamically modulates decoder features according to the provided textual guidance. Central to our approach is the Dynamic Image-Text Alignment (DyITA) module. Rather than simple post-hoc fusion, DyITA integrates text guidance directly into the decoder blocks via a novel dynamic routing mechanism, allowing the model to "switch" its attention patterns based on the requested anatomy and view.
Our main contributions are: \ding{202} We present P2Echo, a text-prompted unified segmentation framework for multi-view echocardiography. It replaces multiple specialist models with a single network that is explicitly controllable via structured textual prompts. \ding{203} We introduce the DyITA module. By employing differentiable routing and token differential operators, DyITA dynamically aligns image features with text prompts inside the decoder, effectively mitigating domain shifts between views. \ding{204} We demonstrate that P2Echo achieves strong performance on a large-scale internal dataset and competitive-to-superior held-out results on three external datasets at substantially lower computational cost, matching or exceeding both specialist models and recent foundation-model adaptations.

\section{Method}
\label{sec:method}

\subsection{Model Architecture}
\begin{figure*}[t]
    \centering
    \includegraphics[width=1\linewidth]{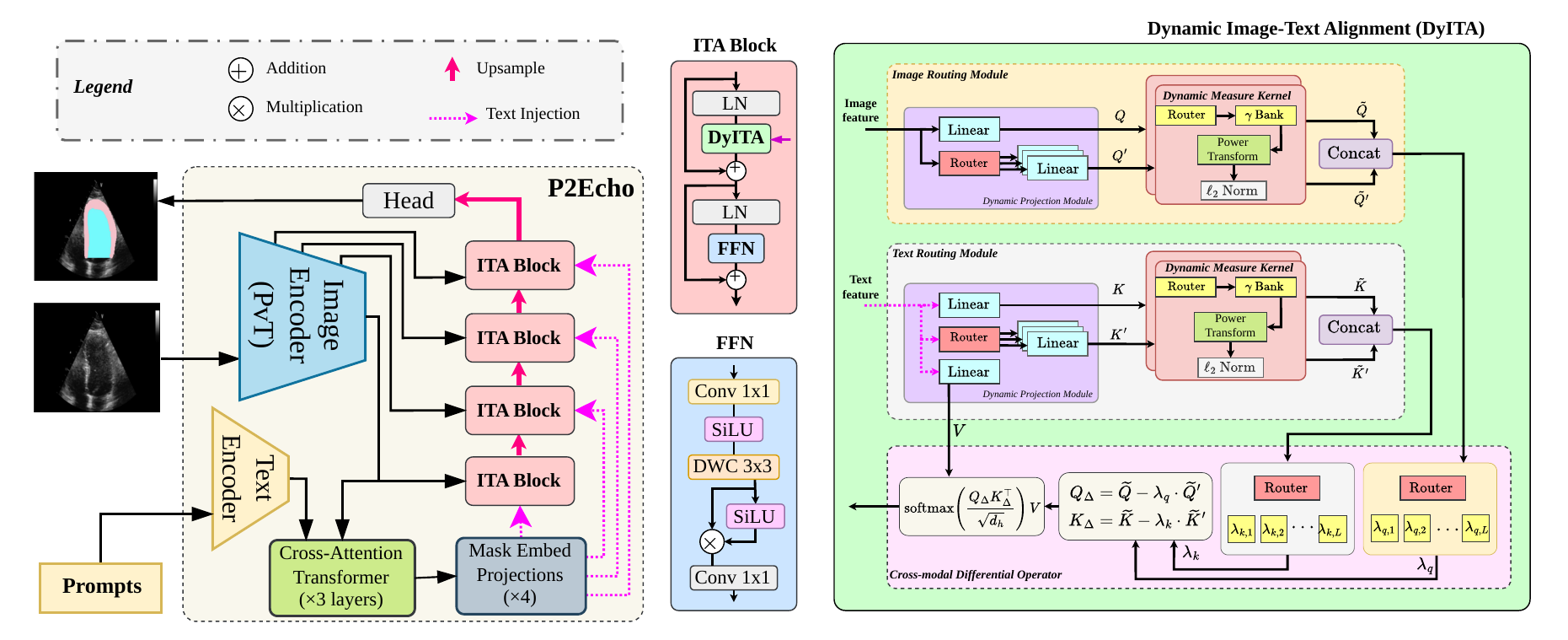}
        \caption{The P2Echo framework. A dual-encoder extracts hierarchical visual and textual features, aligned by a prompt decoder into stage-specific mask embeddings, which are natively fused via ITA blocks.} 
    \label{fig:methodology}
\end{figure*}

P2Echo enables unified multi-view segmentation by conditioning a single network on view-aware text prompts. As illustrated in \cref{fig:methodology}, the architecture comprises three stages: (i) a dual-encoder backbone extracting multi-scale image features and prompt embeddings; (ii) a prompt decoder that grounds prompts in the visual bottleneck to yield image-conditioned mask embeddings; and (iii) a multi-stage decoder natively fusing these embeddings at every spatial resolution via Image-Text Alignment (ITA) blocks.

\subsection{Dual Encoders and Prompt Grounding}
\noindent\textbf{Encoders \& Prompts.} The visual backbone employs PVTv2-B2~\cite{wang2022pvt} to extract a multi-scale feature pyramid $\{F_s\}_{s=1}^{4}$, forwarding shallow features via skip connections to preserve fine spatial details. Concurrently, textual prompts are constructed using a fixed template: \texttt{"View: } $v$\texttt{. Target: }
$k$\texttt{. Description: } $\delta(v,k)$\texttt{."}, where $\delta(v,k)$ is the anatomical description of structure $k$ in view $v$, explicitly encoding view-dependent visibility (\Cref{tab:prompts}). A frozen text encoder embeds these prompts into representations $E \in \mathbb{R}^{B \times N \times D_t}$, which are linearly projected to a shared dimension $d$; here $B$, $N$, and $D_t$ denote the batch size, the
number of requested target structures, and the text embedding dimension,
respectively.

\noindent\textbf{Prompt Decoder.} To bridge the modality gap, prompt tokens are
conditioned on the deepest image feature $F_4$. A transformer-based prompt
decoder (the Cross-Attention Transformer in \cref{fig:methodology}) maps the
prompt embeddings and visual tokens into image-conditioned mask embeddings
$M \in \mathbb{R}^{B \times N \times d}$. These are subsequently projected into
stage-specific mask embeddings $M^{(s)} \in \mathbb{R}^{B \times N \times C_s}$,
matching the channel width $C_s$ of the corresponding decoder stage, to guide
the upsampling pathway. This grounding step establishes the initial image-text correspondence, which ITA blocks subsequently refine at each decoder stage.

\begin{table}[t]
\centering
\caption{View-aware anatomical descriptions used to construct the structured prompts.}
\label{tab:prompts}
{\scriptsize
\setlength{\tabcolsep}{2pt}
\renewcommand{\arraystretch}{1.02}
\begin{tabularx}{\columnwidth}{@{}cX@{}}
\toprule
\textbf{View} & \textbf{Structures and descriptions} \\
\midrule
4CH &
\textbf{LV}: Bullet shaped, forms apex;
\textbf{LA}: posterior to LV;
\textbf{RA}: adjacent to LA, inferior to RV;
\textbf{RV}: triangular, smaller than LV;
\textbf{MYO}: inferoseptal, anterolateral, and apical segments. \\

2CH &
\textbf{LV}: base to apex;
\textbf{LA}: superior to LV;
\textbf{MYO}: anterior and inferior walls. \\

3CH &
\textbf{LV}: LV and LVOT in long axis;
\textbf{LA}: visible in long axis;
\textbf{MYO}: anteroseptal, inferolateral, and apical segments. \\

PSAX &
\textbf{LV}: circular at mid ventricle;
\textbf{RV}: crescent shaped around LV;
\textbf{MYO}: six radial segments. \\
\bottomrule
\end{tabularx}
}
\end{table}

\subsection{Multi-Stage Vision-Language Fusion via DyITA}
At each decoding stage, prompt information is injected through an ITA block to refine cross-modal alignment. Given a stage feature $Y \in \mathbb{R}^{B \times C \times H \times W}$ and mask embeddings $M^{(s)}$, where $C$, $H$, and $W$ denote the stage channel dimension, height, and width, ITA sequentially applies two residual sub-networks: the DyITA module, where prompt semantics explicitly guide the visual features, followed by a Gated Convolutional Feed-Forward Network (FFN) for localized spatial refinement.
\noindent
Inspired by the dynamic differential attention of~\cite{cao2026dynamic}, but unlike its unimodal, linear-attention formulation, DyITA routes flattened image tokens $T \in \mathbb{R}^{B \times S \times C}$ ($S = H \cdot W$) and mask embeddings $M^{(s)}$ through separate redundancy banks and applies Softmax cross-attention at every decoder resolution, via three cascaded operations:

\noindent\textbf{\ding{202} Image and Text Routing Modules (Dynamic Projection):}
DyITA constructs standard projections alongside dynamically routed redundancy
projections. Fixed learned linear mappings produce queries $Q$ from image tokens
$T$, and keys and values $(K,V)$ from the stage-specific prompt embeddings
$M^{(s)}$. In parallel, separate image- and text-side linear routers generate
token-wise assignments for the redundancy queries $Q'$ and redundancy keys
$K'$, respectively. Each router employs a Straight-Through Estimator (STE)
\cite{bengio2013estimating} to perform hard one-hot argmax routing in the
forward pass while using softmax-based surrogate gradients during
backpropagation. The resulting assignments select a specialized projector from
a bank of $P=3$ learned projectors:
\begin{equation}
\begin{aligned}
Q'_i &= \sum_{p=1}^{P} \pi_{i,p}
       \left(T_i W_{p}^{q'}\right), \\
K'_j &= \sum_{p=1}^{P} \rho_{j,p}
       \left(M_{j}^{(s)} W_{p}^{k'}\right),
\qquad P=3,
\end{aligned}
\label{eq:dynamic_projection}
\end{equation}
where $i$ and $j$ index spatial image tokens and prompt tokens, respectively,
while $\pi_{i,p}$ and $\rho_{j,p}$ denote the components of their one-hot
routing assignments. Thus, only one projector contributes to $Q'_i$ or $K'_j$
during the forward pass, while the STE enables end-to-end optimization of the
routers. The separate image- and text-side projector banks allow the two
modalities to learn specialized redundancy representations before cross-modal
differential alignment.

\noindent\textbf{\ding{183} Dynamic Measure Kernel ($\phi_\gamma$):} To adaptively calibrate the projected representations before cross-modal
alignment, four independent Dynamic Measure Kernels are applied to $Q$, $K$, $Q'$, and $K'$, producing $\widetilde{Q}_i$, $\widetilde{K}_j$,
$\widetilde{Q}'_i$, and $\widetilde{K}'_j$, respectively.
Within each kernel, a linear router operates on each
head-specific token vector $z$ and uses an STE-based hard assignment to select
one exponent from an independent learnable bank of nine factors $\{\gamma_1,\ldots,\gamma_9\}$. The factors are initialized to $3.0$, and the
selected token-specific exponent $\gamma(z)$ is constrained to satisfy
$\gamma(z)\geq 1$. The Softplus activation function~\cite{dugas2000incorporating} is applied element-wise to $z$ before exponentiation. The selected exponent then parameterizes the following
norm-preserving power transformation:
\begin{equation}
\phi_{\gamma}(z)
=
\frac{
    \operatorname{Softplus}(z)^{\gamma(z)}
}{
    \left\|
        \operatorname{Softplus}(z)^{\gamma(z)}
    \right\|_2
}
\left\|
    \operatorname{Softplus}(z)
\right\|_2.
\label{eq:dynamic_measure_kernel}
\end{equation}
The kernel is applied independently to each token, e.g.
$\widetilde{Q}_i = \phi_\gamma(Q_i)$ and $\widetilde{Q}'_i = \phi_\gamma(Q'_i)$.
All power operations are applied element-wise. The normalization removes the
magnitude change introduced by exponentiation, while the final scaling restores
the pre-power Softplus norm; hence,
$\|\phi_{\gamma}(z)\|_2
=
\|\operatorname{Softplus}(z)\|_2$.
Accordingly, the kernel modifies the direction and relative component
distribution of each token without changing the magnitude of its
Softplus-transformed representation. This token-adaptive sharpening emphasizes
dominant feature dimensions before the subsequent cross-modal differential
operation.

\noindent\textbf{\ding{204} Cross-Modal Differential Operator:}
To suppress redundancy-driven cross-modal correlations, two linear routers
operate independently on the concatenated
head-specific representations
$[\widetilde{Q};\widetilde{Q}']$ and
$[\widetilde{K};\widetilde{K}']$. Using STE-based hard routing, they select
token-specific subtraction factors $\lambda^{Q}$ and $\lambda^{K}$ from a
shared learnable factor bank
$\{\lambda_p\}_{p=1}^{L}$, where $L=9$ and all factors are initialized
to $0.01$. Thus, for image token $i$ and prompt token $j$,
\begin{equation}
\lambda^{Q}_{i}
=
\sum_{p=1}^{L} r^{Q}_{i,p}\lambda_p,
\qquad
\lambda^{K}_{j}
=
\sum_{p=1}^{L} r^{K}_{j,p}\lambda_p,
\end{equation}
where $r^{Q}_{i}$ and $r^{K}_{j}$ are the corresponding one-hot routing
assignments. The routed redundancy components are then subtracted from the
primary representations:
\begin{equation}
Q_{\Delta,i} = \widetilde{Q}_i - \lambda^{Q}_{i}\,\widetilde{Q}'_i,
\qquad
K_{\Delta,j} = \widetilde{K}_j - \lambda^{K}_{j}\,\widetilde{K}'_j .
\end{equation}
where $Q_\Delta \in \mathbb{R}^{S \times d_h}$ and
$K_\Delta \in \mathbb{R}^{N \times d_h}$ denote the row-wise stacking of
$\{Q_{\Delta,i}\}_{i=1}^{S}$ and $\{K_{\Delta,j}\}_{j=1}^{N}$. This differential formulation is designed to retain prompt-consistent evidence while attenuating redundancy-driven responses.

\noindent\textbf{Softmax Attention \& Prediction.} Cross-modal alignment is finalized using Softmax attention over the $N$ prompt tokens, computed as $O = \mathrm{softmax}\!\left(\frac{Q_{\Delta}K_{\Delta}^{\top}}{\sqrt{d_h}}\right)V$, where $d_h$ is the per-head dimension.
Finally, DyITA is embedded within a residual ITA block and followed by a Layer Normalization (LN) and FFN. The FFN combines $1\times1$ projection, SiLU-activated depth-wise $3\times3$ convolution (DWC), and
element-wise gating to refine local spatial details. The decoder maps these features to $N$ per-pixel logit channels, and at inference, P2Echo is run once with the desired target structure prompt(s), after which argmax over the output channels yields the final segmentation mask. In summary, by repeatedly modulating visual features with the view- and structure-conditioned prompt at every spatial resolution, P2Echo produces view-conditioned representations for multi-plane segmentation within a single network.

\begin{table}[t]
\centering
\caption{Summary of the datasets used for internal development and held-out external evaluation. Entries report the number of annotated samples available for each echocardiographic view, while parenthetical indices denote the corresponding labels in the unified annotation space: 1: LV, 2: MYO, 3: LA, 4: RV, and 5: RA.}
\label{tab:dataset_summary}

\scriptsize
\setlength{\tabcolsep}{3pt}

\begin{tabular}{@{}lcccc@{}}
\toprule
\textbf{Dataset} & \textbf{2CH} & \textbf{3CH} & \textbf{4CH} & \textbf{PSAX} \\
\midrule

\multicolumn{5}{l}{\textit{Internal development}}\\
Camus \cite{leclerc2019deep}            
& 10268\ann{1,2,3} & \na & 10964\ann{1,2,3} & \na \\

CardiacUDA \cite{yang2023graphecho}     
& \na & \na & 1834\ann{1,3,4,5} & \na \\

EchoCP \cite{wang2021echocp}            
& \na & \na & 548\ann{1,3,4,5} & \na \\

EchoNet-Pediatric \cite{reddy2023video}  
& \na & \na & 6423\ann{1} & 8948\ann{1} \\

SegRWMA \cite{liu2023enhance}           
& 941\ann{1} & 1053\ann{1} & 1116\ann{1} & \na \\

\midrule

\multicolumn{5}{l}{\textit{External evaluation}}\\
CardiacNet \cite{yang2024cardiacnet}    
& \na & \na & 2208\ann{1,3,4,5} & \na \\

EchoNet-Dynamic \cite{ouyang2020video}  
& \na & \na & 2554\ann{1} & \na \\

HMC\_QU \cite{degerli2021early}         
& \na & \na & 2283\ann{1,2} & \na \\

\bottomrule
\end{tabular}
\end{table}

\section{Experiments and Results}

\subsection{Dataset Preparation}
\label{sec:datasets_impl}
\begin{figure}[!t]
    \centering
    \includegraphics[width=\columnwidth]{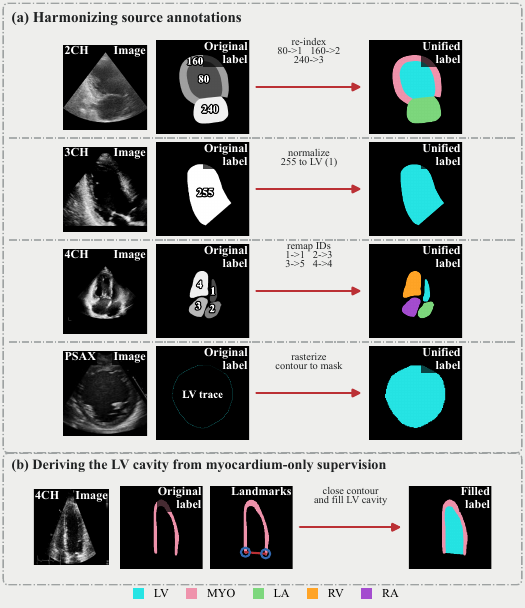}
    \caption{Preprocessing and annotation harmonization for the curated multi-view echocardiography dataset. Colors indicate the corresponding cardiac structures.}
    \label{fig:preprocessing}
\end{figure}

We curate eight echocardiography datasets from multiple sources and views, as summarized in \cref{tab:dataset_summary}. To construct a unified multi-view training set, we first harmonize the heterogeneous annotation protocols used across the source datasets. As illustrated in \cref{fig:preprocessing} (a), source annotations may be represented using dataset-specific categorical pixel values, binary masks, or contour coordinates. We convert all annotations into a common categorical label space, with background assigned to 0 and the cardiac structures indexed as 1 for Left Ventricle (LV), 2 for Myocardium (MYO), 3 for Left Atrium (LA), 4 for Right Ventricle (RV), and 5 for Right Atrium (RA). Dataset-specific pixel values are deterministically remapped to the corresponding unified class identifiers, while contour-based annotations are rasterized and filled to obtain dense segmentation masks. This harmonization enables samples originating from different annotation conventions to be jointly optimized within a single multi-class segmentation framework.
Some source annotations require additional processing before they can be represented in this common label space. In particular, HMC\_QU \cite{degerli2021early} provides myocardial annotations without an explicit LV cavity mask. Following the annotation harmonization strategy of EchoONE \cite{hu2025echoone}, we identify the basal endpoints of the myocardial annotation, close the open contour between these landmarks, and fill the resulting enclosed region to derive the LV cavity while retaining the original MYO annotation, as illustrated in \cref{fig:preprocessing}(b). After harmonization, all datasets share a consistent semantic representation despite differences in their original annotation formats. The resulting unified masks are subsequently used for training and evaluation across the supported echocardiographic views.

\begin{table*}[h]
\centering
\caption{Comparison of segmentation methods across different echocardiography views. Values are reported as mean $\pm$ standard deviation. The best and second-best results are highlighted in \textcolor{blue}{blue} and \textcolor{red}{red}, respectively. Statistical significance compared with the second-best method was evaluated using the Wilcoxon signed-rank test; ** indicates $p < 0.01$. P2Echo consistently achieves superior segmentation performance and improved robustness across most echocardiographic views.}
\label{tab:comparison}
\setlength{\tabcolsep}{3.2pt}
\renewcommand{\arraystretch}{1.1}
\resizebox{\textwidth}{!}{%
\begin{tabular}{l|ccc|ccc|ccc|ccc}
\toprule
\multirow{2}{*}{Method} & \multicolumn{3}{c|}{2CH} & \multicolumn{3}{c|}{3CH} & \multicolumn{3}{c|}{4CH} & \multicolumn{3}{c}{PSAX} \\
& mDice & mIoU & HD95 & mDice & mIoU & HD95 & mDice & mIoU & HD95 & mDice & mIoU & HD95 \\
\midrule
nnUNet~\cite{isensee2021nnu} &
$92.82 \pm 3.75$ & $86.82 \pm 6.14$ & $4.90 \pm 2.20$ &
$93.07 \pm 2.97$ & $87.17 \pm 5.03$ & $5.98 \pm 3.02$ &
$\second{91.39} \pm 5.68$ & $84.64 \pm 8.17$ & $6.35 \pm 4.02$ &
$91.50 \pm 5.69$ & $84.78 \pm 8.54$ & $7.72 \pm 4.86$ \\
SwinUNet~\cite{cao2022swin} &
$86.58 \pm 6.08$ & $77.15 \pm 7.93$ & $7.75 \pm 3.76$ &
$87.27 \pm 6.03$ & $77.88 \pm 8.89$ & $8.86 \pm 4.97$ &
$85.43 \pm 8.16$ & $75.62 \pm 10.04$ & $9.04 \pm 6.51$ &
$85.29 \pm 8.96$ & $75.29 \pm 12.06$ & $11.11 \pm 7.10$ \\
ResUNet~\cite{zhang2018road} &
$91.80 \pm 2.89$ & $85.21 \pm 4.53$ & $5.46 \pm 3.30$ &
$91.78 \pm 3.60$ & $85.26 \pm 5.81$ & $5.69 \pm 4.01$ &
$88.09 \pm 6.52$ & $80.01 \pm 8.57$ & $7.83 \pm 10.28$ &
$88.99 \pm 7.45$ & $80.86 \pm 10.67$ & $14.88 \pm 15.12$ \\
MissFormer~\cite{huang2022missformer} &
$90.88 \pm 3.04$ & $83.70 \pm 4.73$ & $5.48 \pm 1.62$ &
$91.09 \pm 5.65$ & $84.12 \pm 8.43$ & $5.95 \pm 4.07$ &
$88.34 \pm 5.67$ & $79.92 \pm 7.72$ & $6.24 \pm 4.16$ &
$88.50 \pm 7.24$ & $80.00 \pm 10.26$ & $9.20 \pm 5.19$ \\
CENet \cite{bozorgpour2025cenet} &
$93.11 \pm 2.14$ & $87.38 \pm 3.60$ & $4.07 \pm 1.25$ &
$93.09 \pm 2.99$ & $87.42 \pm 5.03$ & $4.42 \pm 2.10$ &
$90.88 \pm 4.80$ & $83.89 \pm 6.56$ & $4.71 \pm 3.02$ &
$91.08 \pm 5.10$ & $83.99 \pm 8.02$ & $\second{7.29} \pm 3.46$ \\
\midrule
SAMUS~\cite{lin2024beyond} &
$89.04 \pm 10.39$ & $81.85 \pm 12.13$ & $13.67 \pm 3.38$ &
$\second{93.61} \pm 2.91$ & $88.13 \pm 4.97$ & $8.42 \pm 1.18$ &
$89.36 \pm 9.15$ & $82.00 \pm 11.44$ & $13.72 \pm 3.47$ &
$91.46 \pm 4.23$ & $84.55 \pm 6.86$ & $12.28 \pm 2.17$ \\
SAM3 (Zero-shot)~\cite{carion2025sam}
& $8.75 \pm 6.41$ & $4.84 \pm 4.23$ & $220.04 \pm 7.70$
& $2.37 \pm 1.34$ & $1.29 \pm 0.26$ & $310.12 \pm 20.45$
& $6.32 \pm 4.22$ & $3.53 \pm 2.01$ & $241.67 \pm 30.12$
& $5.06 \pm 2.66$ & $2.99 \pm 2.20$ & $306.86 \pm 11.85$ \\
SAM3 (Fine-tuned)~\cite{carion2025sam} &
$\second{93.39} \pm 1.75$ & $\second{87.60} \pm 3.11$ & $3.18 \pm 0.89$ &
$92.70 \pm 2.48$ & $87.04 \pm 4.21$ & $6.69 \pm 1.70$ &
$90.85 \pm 5.25$ & $83.85 \pm 7.59$ & $4.19 \pm 3.53$ &
$91.18 \pm 4.66$ & $84.13 \pm 7.32$ & $7.90 \pm 3.85$ \\
MedCLIP-SAMv2 \cite{koleilat2025medclip} & $28.98 \pm 16.00$ & $17.98 \pm 11.14$ & $170.40 \pm 79.67$ & $31.70 \pm 12.76$ & $19.48 \pm 8.57$ & $170.09 \pm 60.78$ & $24.61 \pm 14.78$ & $14.87 \pm 10.07$ & $170.90 \pm 86.40$ & $27.72 \pm 15.34$ &
  $17.04 \pm 10.72$ & $36.48 \pm 10.60$ \\
Grounding DINO-US-SAM \cite{rasaee2025grounding} &
$73.97 \pm 21.04$ & $62.20 \pm 21.16$ & $15.47 \pm 19.39$ &
$83.29 \pm 7.09$ & $71.92 \pm 9.60$ & $10.59 \pm 5.61$ &
$76.43 \pm 13.88$ & $64.30 \pm 16.02$ & $12.31 \pm 7.41$ &
$84.42 \pm 9.54$ & $74.06 \pm 12.55$ & $13.50 \pm 8.19$ \\
EchoONE~\cite{hu2025echoone} &
$91.69 \pm 11.67$ & $85.83 \pm 12.64$ & $\second{2.51} \pm 3.38$ &
$93.60 \pm 3.35$ & $\second{88.15} \pm 5.48$ & $\second{4.24} \pm 2.19$ &
$91.11 \pm 10.10$ & $\second{84.69} \pm 11.62$ & $\second{3.88} \pm 3.50$ &
$\second{91.71} \pm 8.82$ & $\best{85.56} \pm 10.70$ & $7.97 \pm 3.06$ \\
\midrule
P2Echo (Ours) &
$\best{96.94}{**} \pm 1.72$ & $\best{94.11}{**} \pm 3.08$ & $\best{2.03}{**} \pm 0.92$ &
$\best{94.71}{**} \pm 2.16$ & $\best{90.03}{**} \pm 3.79$ & $\best{3.96}{**} \pm 1.60$ &
$\best{93.87}{**} \pm 5.07$ & $\best{88.81}{**} \pm 7.48$ & $\best{3.33}{**} \pm 3.77$ &
$\best{92.00} \pm 4.97$ & $\second{85.52} \pm 7.76$ & $\best{6.99} \pm 3.74$ \\ \midrule
\end{tabular}%
}
\end{table*}

\subsection{Implementation Details}
We train and internally evaluate on five datasets and reserve three datasets for held-out external evaluation. Each development dataset is split into $80\%$ training, $10\%$ validation, and $10\%$ internal test data. P2Echo is implemented in PyTorch with $256\times256$ inputs and an ImageNet-pretrained PVTv2 B2 visual encoder \cite{wang2022pvt}. The model is trained for 200 epochs using SGD with a batch size of 32, an initial learning rate of $0.01$, momentum of $0.9$, and weight decay of $10^{-4}$. The learning rate follows a polynomial decay schedule with power $0.9$, and gradients are clipped to a maximum norm of $1.0$. We employ mixed precision training and standard spatial and intensity augmentations. The checkpoint achieving the highest mean validation Dice is retained for evaluation. Text prompts are encoded using the frozen Qwen3-Embedding-0.6B model \cite{zhang2025qwen3}, the smallest member of the Qwen3 embedding family, chosen so that the language pathway contributes negligibly to inference cost; as the encoder is frozen and the prompt set is fixed by the view–structure template, its embeddings are computed once and reused. Identical view-aware prompts are used across all text-based comparison methods. We optimize P2Echo using the Boundary Difference over Union (BDoU) loss \cite{sun2023boundary}, which extends region overlap optimization by placing greater emphasis on mismatched regions near object boundaries.

Given a predicted mask $\hat{Y}$ and its corresponding ground truth $G$, the BDoU loss is formulated as

\begin{equation}
\mathcal{L}_{\text{BDoU}} =
\frac{|G \cup \hat{Y}| - |G \cap \hat{Y}|}{|G \cup \hat{Y}| - \alpha|G \cap \hat{Y}|},
\end{equation}

where $\alpha$ controls the relative emphasis placed on boundary agreement, following \cite{sun2023boundary}; $\alpha$ is adaptively determined from the ratio between the target boundary and its foreground area. We additionally apply deep supervision to the main prediction and two intermediate decoder outputs, encouraging accurate segmentation throughout the decoding process.

\subsection{Evaluation Metrics}

Segmentation performance is evaluated using the Dice Similarity Coefficient (mDice), Intersection over Union (mIoU), and the 95th percentile Hausdorff Distance (HD95). For a predicted binary mask $\hat{Y}$ and its corresponding ground-truth mask $G$, Dice, and IoU are defined as

\begin{equation}
\begin{aligned}
\mathrm{Dice}(\hat{Y},G) = \frac{2|\hat{Y} \cap G|}{|\hat{Y}| + |G|}, \qquad
\mathrm{IoU}(\hat{Y},G) = \frac{|\hat{Y} \cap G|}{|\hat{Y} \cup G|}.
\end{aligned}
\end{equation}

HD95 measures boundary disagreement and is computed as the 95th percentile of the bidirectional minimum distances between the predicted and ground-truth boundaries. Higher Dice and IoU indicate greater regional overlap, whereas lower HD95 indicates more accurate boundary localization. It is noteworthy that Dice scores of $0$ from failed class predictions are retained in the average, whereas infinite HD95 values are excluded. Thus, methods that completely miss a structure, such as MYO on HMC\_QU, are penalized in Dice but not equivalently in HD95, which slightly favors the baselines relative to P2Echo.

\subsection{Results}

\subsubsection{Internal Multi-View Evaluation}
We evaluate the performance of P2Echo on the internal test split and compare it with traditional models such as nnUNet~\cite{isensee2021nnu}, SAM-based models such as SAMUS~\cite{lin2024beyond} and EchoONE~\cite{hu2025echoone}, and text-guided approaches such as SAM3~\cite {carion2025sam}, MedCLIP-SAMv2~\cite{koleilat2025medclip}, and Grounding DINO-US-SAM \cite{rasaee2025grounding}; see \cref{tab:comparison}. P2Echo achieves strong segmentation accuracy across all views, with improved Dice, IoU, and HD95 on most views. Compared to EchoONE \cite{hu2025echoone}, the only prior work targeting multi-view echo segmentation, P2Echo improves on most views (+5.3 Dice on 2CH, +1.1 on 3CH, +2.8 on 4CH; comparable on PSAX), using a single set of parameters rather than per-plane weights or learnable dense prompts. This is consistent with view- and structure-conditioned prompting providing an effective, explicitly specifiable alternative to per-view specialization.

\subsubsection{External Held-Out Generalization}
To assess the generalization and robustness to domain shifts from different scanners and acquisition protocols, we evaluate P2Echo on three external datasets in a held-out setting. As shown in \Cref{tab:external}, P2Echo maintains high performance and outperforms text-based baselines such as SAM3 \cite{carion2025sam}, MedCLIP-SAMv2~\cite{koleilat2025medclip}, and Grounding DINO-US-SAM \cite{rasaee2025grounding}. The substantially lower performance of generic foundation model baselines indicates limited transfer to the echocardiographic domain without task-specific adaptation. In contrast, P2Echo maintains structural coherence across multi-source settings.
\begin{figure*}[h]
    \centering
    \includegraphics[width=0.9\linewidth]{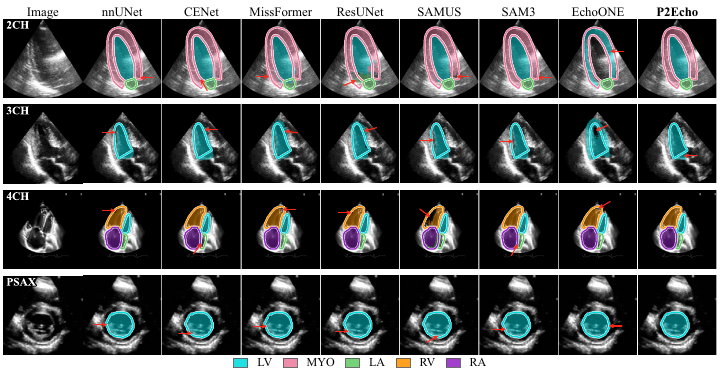}
    \caption{Qualitative comparison across four echocardiographic views (rows). Columns show the input image, baseline predictions, and our P2Echo. Colored regions denote predictions, while class-colored contours indicate ground-truth (GT) boundaries. Red arrows highlight representative segmentation errors discussed in the text.}
    \label{fig:qualitative_results}
\end{figure*}

\subsubsection{Qualitative Analysis}
We further inspect the qualitative results of our method. As illustrated in \cref{fig:qualitative_results}, baseline methods struggle with less consistently represented structures, such as myocardium and right-sided chambers, producing myocardial boundary errors in 2CH and incomplete right-ventricular or right-atrial masks in 4CH. In particular, EchoONE confuses the LV cavity with the myocardial wall in the shown 2CH case, with multiclass fusion further degrading myocardial delineation, and leaves a hole near the LV apex in 3CH. Residual cavity boundary errors are also visible in PSAX. In contrast, P2Echo delineates cavities more consistently across these examples, including challenging apical cases, although mild oversegmentation remains at the basal LV boundary in 3CH. This behavior aligns with the view-aware prompts, which provide view-specific shape and location cues that help guide segmentation in low-contrast or signal-dropout regions. \Cref{fig:p2echo_qual_all_rows} (a) demonstrates P2Echo's prompt-driven flexibility by generating anatomically plausible masks for targets that are not annotated in the available ground truth. Finally, to verify the effect of DyITA, we visualize feature maps before and after applying the DyITA block in \cref{fig:p2echo_qual_all_rows} (b). The maps before DyITA are diffuse and not well localized to the target region. After DyITA, activations are more concentrated on the prompted anatomy, which is consistent with sharper boundaries and improved contouring in the final predictions.

\subsubsection{Ablation Study}
\begin{figure*}[!h]
    \centering
    \includegraphics[width=0.9\linewidth]{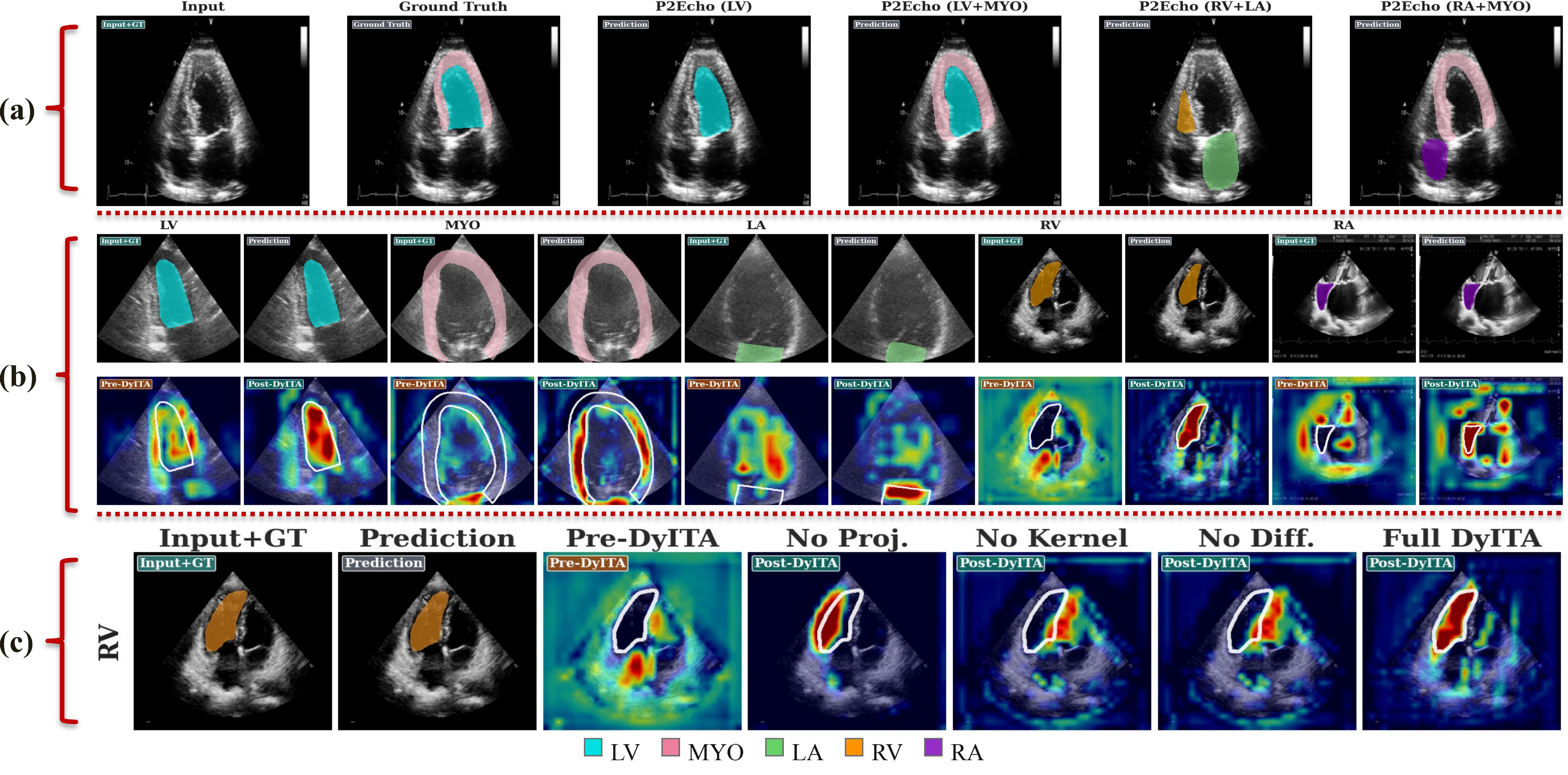}
    \caption{Qualitative analysis of P2Echo. (a) Prompt-driven segmentation of structures beyond available annotations. (b) Feature activation maps before and after DyITA for different target structures. (c) Visualization of DyITA component ablations.}
    \label{fig:p2echo_qual_all_rows}
\end{figure*}

We evaluate the contribution of the three DyITA sub-modules through component-wise ablations on HMC\_QU (\Cref{tab:ablation}). Replacing the dynamic projection with a static linear layer, removing the dynamic measure kernel, and removing the cross-modal differential operator cost 2.18, 1.70, and 2.19 mDice and add 0.39, 0.43, and 0.92 HD95. Removing any one of the three degrades both metrics, and the differential operator accounts for most of the boundary gain, matching its role of suppressing redundancy-driven cross-modal correlations before attention. This decline is backed up visually (\cref{fig:p2echo_qual_all_rows} (c)): ablated variants attend more diffusely, which causes noisy activations that bleed into adjacent structures and blur target boundaries, whereas the full DyITA focuses sharply on the intended anatomy. Finally, increasing the text encoder from 0.6B to 4B yields only a 0.07-point mDice improvement while slightly increasing HD95, supporting the lightweight 0.6B encoder used in P2Echo.

\begin{table*}[!h]
\centering
\caption{Performance comparison on external datasets. FLOPs are reported in (G), and the number of parameters is reported in (M). Values are presented as mean $\pm$ standard deviation. Statistical significance compared with the second-best method was evaluated using the Wilcoxon signed-rank test; ** indicates $p < 0.01$. P2Echo achieves superior generalization across diverse external echocardiography datasets, providing competitive segmentation accuracy with reduced computational cost and improved performance stability compared with existing methods.}
\label{tab:external}
  \resizebox{\textwidth}{!}{
  \setlength{\tabcolsep}{1.5pt} 
  \begin{tabular}{@{}l c c cccccc@{}}
    \toprule
    \multirow{2}{*}{Method}
    & \multirow{2}{*}{Params (M)} 
    & \multirow{2}{*}{FLOPs (G)}
    & \multicolumn{2}{c}{HMC\_QU}
    & \multicolumn{2}{c}{EchoNet-Dynamic}
    & \multicolumn{2}{c}{CardiacNet} \\
    \cmidrule(lr){4-5} \cmidrule(lr){6-7} \cmidrule(lr){8-9}
     & & & mDice & HD95 & mDice & HD95 & mDice & HD95 \\
    \midrule
    nnUNet~\cite{isensee2021nnu} & 33.48 & 28.30 &
      $45.30 \pm 45.41$ & $\second{7.64} \pm 3.78$ &
      $84.78 \pm 7.40$ & $15.08 \pm 8.09$ &
      $88.13 \pm 10.98$ & $6.32 \pm 6.27$ \\
    SwinUNet~\cite{cao2022swin} & 27.17 & 6.16 &
      $52.24 \pm 11.51$ & $19.28 \pm 15.53$ &
      $82.15 \pm 9.47$ & $16.06 \pm 9.22$ &
      $75.42 \pm 14.52$ & $11.40 \pm 7.45$ \\
    ResUNet~\cite{zhang2018road} & 13.05 & 54.54 &
      $58.02 \pm 12.26$ & $23.04 \pm 14.05$ &
      $78.19 \pm 11.00$ & $34.50 \pm 19.63$ &
      $86.82 \pm 14.30$ & $6.22 \pm 7.38$ \\
    MissFormer~\cite{huang2022missformer} & 42.46 & 9.89 &
      $48.31 \pm 11.80$ & $23.24 \pm 12.21$ &
      $85.37 \pm 7.43$ & $12.82 \pm 8.13$ &
      $86.07 \pm 13.70$ & $5.98 \pm 6.75$ \\
    CENet \cite{bozorgpour2025cenet} & 33.39 & 12.76 &
      $54.41 \pm 13.25$ & $19.89 \pm 12.65$ &
      $85.67 \pm 6.91$ & $\best{11.33} \pm 6.54$ &
      $89.56 \pm 13.48$ & $4.33 \pm 6.68$ \\
    \midrule
    SAMUS~\cite{lin2024beyond} & 39.65 & 145.88
    & $\second{61.93} \pm 27.99$ & $19.19 \pm 14.77$
    & $85.58 \pm 7.02$ & $18.63 \pm 4.61$
    & $86.64 \pm 10.92$ & $8.94 \pm 5.48$ \\
    SAM3 (Zero-shot)~\cite{carion2025sam} & - & 2417.08
    & $1.91 \pm 1.00$ & $261.99 \pm 3.54$
    & $3.25 \pm 2.05$ & $192.50 \pm 2.14$
    & $2.14 \pm 2.00$ & $173.12 \pm 7.95$ \\
    SAM3 (Fine-tuned)~\cite{carion2025sam} & 14.40 & 2417.08 &
      $43.46 \pm 43.43$ & $179.59 \pm 32.25$&
      $85.27 \pm 7.45$ & $15.58 \pm 7.21 $&
      $\second{90.62} \pm 9.84$ & $\second{3.70} \pm 5.38$ \\
    MedCLIP-SAMv2~\cite{koleilat2025medclip} & 836.99 & 2458.17 &
      $24.25 \pm 14.73$ & $78.30 \pm 27.05$ &
      $30.57 \pm 21.93$ & $35.84 \pm 12.17$ &
      $16.20 \pm 8.24$ & $232.61 \pm 57.47$ \\
    Grounding DINO-US-SAM \cite{rasaee2025grounding} & 397.29 & 1171.38 &
      $60.99 \pm 28.08$ & $13.13 \pm 6.31$ &
      $81.74 \pm 6.78$ & $16.53 \pm 6.98$ &
      $78.05 \pm 21.81$ & $9.92 \pm 9.67$ \\
    EchoONE~\cite{hu2025echoone} & 45.62 & 174.34
    & $53.96 \pm 31.43$ & $13.10 \pm 16.69$
    & $\second{86.45} \pm 5.74$ & $15.38 \pm 4.30$
    & $88.10 \pm 10.43$ & $3.84 \pm 5.24$ \\
    \midrule
    P2Echo (Ours) & 37.47 & 6.90 &
      $\best{82.22}{**} \pm 9.53$ & $\best{6.92}{**} \pm 2.58$ &
      $\best{87.79}{**} \pm 7.01$ & $\second{11.89} \pm 6.97$ &
      $\best{92.95}{**} \pm 10.02$ & $\best{3.37}{**} \pm 5.63$ \\ \midrule

  \end{tabular}
  }
\end{table*}

\begin{table}[t]
\centering
\caption{Text encoder scale and DyITA component ablation.}
\label{tab:ablation}
\footnotesize
\setlength{\tabcolsep}{4pt}
\renewcommand{\arraystretch}{1.1}
\begin{tabular}{@{}l l c c@{}}
\toprule
\textbf{Ablation} & \textbf{Variant} & \textbf{mDice} & \textbf{HD95} \\
\midrule
\multirow{2}{*}{\textbf{Text encoder}}
 & Qwen (0.6B) & 82.22 & 6.92 \\
 & Qwen (4B)   & 82.29 & 6.95 \\
\midrule
\multirow{4}{*}{\textbf{DyITA components}}
 & w/o Dynamic Projection     & 80.04 & 7.31 \\
 & w/o Dynamic Measure Kernel & 80.52 & 7.35 \\
 & w/o Differential Operator  & 80.03 & 7.84 \\
 & Full DyITA                 & \textbf{82.22} & \textbf{6.92} \\
\bottomrule
\end{tabular}
\end{table}

\subsubsection{Effect of Text-Prompt Formulation \& Runtime Analysis}

\Cref{tab:promptint} varies only the prompt text on 450 images from the internal test split (950 annotated structure instances).
The canonical template achieves the highest performance with 93.53\% Dice. 
Reformulating the same semantic request using free clinical language reduces performance, with clinical paraphrases and concise prompts achieving 80.48\% and 35.83\% Dice, respectively. The largest degradation is observed for structures whose identification depends on relational anatomical descriptions rather than simple shape characteristics. These results indicate that the current model remains sensitive to the prompt formulation used during training, highlighting the importance of structured and consistent textual conditioning. The high variance observed for concise prompts (35.83 $\pm$ 36.23) further suggests that robustness to unconstrained clinical language remains challenging. Future work will investigate paraphrase-augmented training and improved language alignment strategies to enhance tolerance to diverse prompt formulations. The gray rows demonstrate controlled abstention behavior for unsupported, ambiguous, and null requests. The main prompt templates were developed with radiologist input to maintain anatomical consistency and clinical appropriateness.

\begin{table*}[t]
\centering
\caption{Prompt intervention on 450 images (950 structure instances).
Only the prompt text is varied. Gray rows probe unsupported requests, where
near-zero overlap is the intended behavior.}
\label{tab:promptint}
\setlength{\tabcolsep}{6pt}
\renewcommand{\arraystretch}{1.15}
\begin{tabular}{llcc}
\toprule
Condition & Example prompt & Dice (\%) $\uparrow$ & IoU (\%) $\uparrow$ \\
\midrule
Canonical & \emph{``Segment the LV: The LV is fully visualized from base to apex.''}
 & \textbf{93.53\,$\pm$\,6.83} & \textbf{88.39\,$\pm$\,8.84} \\
Clinical paraphrase & \emph{``Delineate the left ventricular blood pool.''}
 & 80.48\,$\pm$\,23.88 & 72.62\,$\pm$\,26.76 \\
Concise & \emph{``Segment the left ventricle cavity.''}
 & 35.83\,$\pm$\,36.23 & 28.58\,$\pm$\,30.95 \\
\midrule
\rowcolor{gray!12} Ambiguous & \emph{``Segment the cardiac chamber.''}
 & 0.01\,$\pm$\,0.23 & 0.01\,$\pm$\,0.12 \\
\rowcolor{gray!12} Irrelevant & \emph{``Segment the liver.''}
 & 6.35\,$\pm$\,19.70 & 4.95\,$\pm$\,16.54 \\
\rowcolor{gray!12} Null & \emph{``No anatomical structure is requested.''}
 & 0.00\,$\pm$\,0.00 & 0.00\,$\pm$\,0.00 \\
\bottomrule
\end{tabular}
\end{table*}

\noindent\textbf{Inference efficiency.}
P2Echo processes in $0.01546\pm0.00002$ s/image
(64.68 FPS), ranking third in latency (\cref{tab:runtime}). Only SwinUNet and MissFormer are faster,
but both have considerably lower segmentation accuracy in \Cref{tab:comparison,tab:external}.
P2Echo is $1.51\times$ faster than CENet and $4.28$--$159.36\times$
faster than the evaluated promptable methods. Our approach achieves strong segmentation performance while maintaining high adaptability and computational efficiency. 

\begin{table}[t]
\centering
\caption{Inference time per image on an NVIDIA H100 (FP32, batch size 1; mean $\pm$ standard deviation over 10 runs).}
\label{tab:runtime}
\footnotesize
\renewcommand{\arraystretch}{0.92}
\begin{tabular}{lc}
\toprule
Method & Time (s/image) $\downarrow$ \\
\midrule
nnUNet & 0.04057 $\pm$ 0.00002 \\
SwinUNet & 0.00577 $\pm$ 0.00002 \\
MissFormer & 0.01034 $\pm$ 0.00001 \\
CENet & 0.02328 $\pm$ 0.00010 \\
\midrule
SAMUS & 0.06620 $\pm$ 0.00024 \\
SAM3 (Zero-shot) & 0.19694 $\pm$ 0.00004 \\
SAM3 (Fine-tuned) & 0.08203 $\pm$ 0.00009 \\
MedCLIP-SAMv2 & 2.46376 $\pm$ 0.02072 \\
Grounding DINO-US-SAM & 2.59870 $\pm$ 0.01970 \\
EchoONE & 0.08649 $\pm$ 0.00004 \\
\midrule
\textbf{P2Echo (Ours)} & \textbf{0.01546 $\pm$ 0.00002} \\
\bottomrule
\end{tabular}
\end{table}

\section{Discussion}
\label{sec:discussion}

The results demonstrate that a single text-conditioned model can achieve strong segmentation across heterogeneous echocardiographic views without maintaining separate view-specific networks. P2Echo matches or exceeds the strongest baseline across views, with the largest margin on 2CH and parity on PSAX. On the external datasets with natively annotated structures (EchoNet-Dynamic, CardiacNet), P2Echo shows consistent transfer across acquisition domains. The larger margin observed on HMC\_QU has a specific explanation: MYO supervision is available only from CAMUS during training, limiting the diversity of domains in which this structure is represented. In heterogeneous datasets with incomplete annotation overlap, conventional segmentation models may implicitly associate the absence of a label with the absence of the corresponding anatomy, resulting in biased predictions toward frequently observed structures \cite{zhang2021dodnet,islam2026comparison}. This effect is evident in HMC\_QU, where baseline models show severe MYO degradation (e.g., nnUNet) while maintaining comparable LV segmentation. In contrast, P2Echo preserves more coherent myocardial segmentation, suggesting that explicit structure-aware conditioning helps mitigate annotation-space bias under domain shift. Crucially, P2Echo attains these results at 6.90 GFLOPs and 64.68 FPS, 4–160× more efficient than the promptable baselines, making a single controllable network a practical replacement for multiple view-specific specialists. The ablation results support the contribution of both textual conditioning and DyITA. The prompt intervention demonstrates sensitivity to prompt formulation, highlighting the importance of structured textual conditioning. Removing individual DyITA components consistently reduces performance, confirming the importance of the proposed dynamic alignment strategy. 

Several limitations remain. First, the current experiments focus on four standard echocardiographic views and assume that the view identity is available when constructing the prompt. Therefore, the present results do not establish generalization to previously unseen views. Second, the source datasets contain heterogeneous annotation sets and uneven view coverage. In particular, 3CH and PSAX supervision is available from only one development dataset each, reflecting the limited availability of publicly accessible segmentation datasets for these views. Despite this limited source diversity, P2Echo achieves strong performance on both 3CH and PSAX, suggesting robust learning across the available multi-view supervision. Finally, the current evaluation focuses on segmentation accuracy and not downstream clinical measurements. Future work will extend P2Echo to additional echocardiographic views and investigate whether improved segmentation translates into more accurate and reproducible cardiac measurements and functional assessment.

\section{Conclusion}
\label{sec:conclusion}
We present P2Echo, a unified text-prompted framework for multi-view echocardiography segmentation. By conditioning a single network on structured view- and structure-aware prompts through our DyITA module, P2Echo suppresses cross-modal redundancy and sharpens boundaries while remaining explicitly controllable. Experiments demonstrate strong internal accuracy and competitive external results across diverse datasets at substantially lower computational cost than comparable promptable models, offering an efficient, controllable single-model solution for multi-plane cardiac assessment.

\section*{Acknowledgment}

This work was supported by the Canadian Foundation for Innovation-John R. Evans Leaders Fund (CFI-JELF) program grant number 42816. Mitacs Accelerate program grant number AWD024298-IT33280. We also acknowledge the support of the Natural Sciences and Engineering Research Council of Canada (NSERC), [RGPIN-2023-03575]. Cette recherche a été financée par le Conseil de recherches en sciences naturelles et en génie du Canada (CRSNG), [RGPIN-2023-03575].  

% %\clearpage
\bibliographystyle{IEEEtran}
\bibliography{ref} 
\end{document}